\pdfoutput=1

\documentclass[11pt]{article}
\usepackage[final]{ACL}
\newcommand{\eat}[1]{}

\usepackage{latexsym}
\usepackage{amsfonts}
\usepackage{amsmath}
\usepackage{xcolor}
\usepackage{colortbl}
\usepackage{epsfig}
\usepackage{xspace}
\usepackage{graphicx}
\usepackage{subfigure}
\usepackage{paralist}
\usepackage{enumerate}
\usepackage[color,matrix,arrow,all]{xy}
\usepackage{comment}
\usepackage{booktabs}
\usepackage{balance}
\usepackage{stmaryrd}
\usepackage{pifont}
\usepackage{hhline}
\usepackage{listings}
\usepackage{array}
\usepackage{float}
\usepackage[flushleft]{threeparttable}

\usepackage{mathrsfs}
\usepackage{makecell}
\usepackage{xparse}
\usepackage{wrapfig}

\usepackage{epsfig}
\usepackage{multirow}
\usepackage{url}

\usepackage{multirow}
\usepackage{natbib}
\usepackage{graphicx}

\usepackage{listings}
\usepackage{framed}
\usepackage{xcolor}
\usepackage{color}
\usepackage{geometry}
\usepackage{changepage}
\colorlet{shadecolor}{gray!20}

\definecolor{shadecolor}{RGB}{220,220,220}

\definecolor{inputcolor}{RGB}{255,139,35}
\definecolor{outputcolor}{RGB}{120,212,252}
\definecolor{embedcolor}{RGB}{254,127,156}
\definecolor{maskcolor}{RGB}{122,128,255}
\definecolor{ecolor}{RGB}{58,149,54}

\definecolor{highcolor}{RGB}{255,153,153}
\definecolor{midcolor}{RGB}{255,204,204}
\definecolor{lowcolor}{RGB}{204,229,255}

\usepackage{tikz}
\usetikzlibrary{shapes,decorations}
\usetikzlibrary{calc}

\usepackage[export]{adjustbox}

\definecolor{green}{RGB}{0,128,0}

\definecolor{yellow}{RGB}{255,200,18}

\newcommand{\stab}{\vspace{1.2ex}\noindent}

\newcommand{\bi}{\begin{itemize}}
\newcommand{\ei}{\end{itemize}}

\newcommand{\be}{\begin{enumerate}}
\newcommand{\ee}{\end{enumerate}}
\newcommand{\beqn}{\begin{eqnarray*}}
\newcommand{\eeqn}{\end{eqnarray*}}

\newcommand{\stitle}[1]{\stab\noindent{\bf #1}}

\newcommand{\eg}{{\em e.g.,}\xspace}

\makeatletter
    \newcommand\figcaption{\def\@captype{figure}\caption}
    \newcommand\tabcaption{\def\@captype{table}\caption}
\makeatother

\tikzstyle{mybox} = [draw=black, fill=black!5, thick,
   rectangle, rounded corners, inner sep=0pt, inner ysep=6pt]
\tikzstyle{fancytitle} =[fill=black, text=white]

\NewDocumentCommand{\nan}{ mO{} }{\textcolor{blue}{\textsuperscript{\textit{Nan}}\textsf{\textbf{\small[#1]}}}}

\NewDocumentCommand{\yang}{ mO{} }{\textcolor{green}{\textsuperscript{\textit{yang}}\textsf{\textbf{\small[#1]}}}}

\NewDocumentCommand{\zzx}{ mO{} }{\textcolor{yellow}{\textsuperscript{\textit{zzx}}\textsf{\textbf{\small[#1]}}}}

\usepackage{amsmath}
\usepackage{graphicx}
\usepackage{times}
\usepackage{amssymb}
\usepackage{multirow}

\usepackage[T1]{fontenc}

\usepackage[utf8]{inputenc}

\usepackage{microtype}

\usepackage{inconsolata}

\usepackage[ruled,linesnumbered]{algorithm2e}
\usepackage{booktabs} 
\usepackage{siunitx}

\usepackage{multirow}
\usepackage{mdframed}
\usepackage{amsmath}
\usepackage{graphicx} 
\usepackage{multirow}
\usepackage{xspace}
\usepackage{mdframed}
\newcommand{\ben}{{MEBench}\xspace}
\newcommand{\meqa}{{MEQA}\xspace}
\title{MEBench: Benchmarking Large Language Models for Cross-Document Multi-Entity Question Answering}

\author{
 \textbf{Teng Lin\textsuperscript{1}},
 \textbf{Yuyu Luo\textsuperscript{1,2}},
 \textbf{Honglin Zhang\textsuperscript{3}},
 \textbf{Jicheng Zhang\textsuperscript{3}},
\\ 
 \textbf{Chunlin Liu\textsuperscript{3}},
 \textbf{Kaishun Wu\textsuperscript{1}},
 \textbf{Nan Tang\textsuperscript{1,2}\thanks{Nan Tang is the corresponding author.}}
 \\
  \textsuperscript{1}The Hong Kong University of Science and Technology (Guangzhou)
  \\
 \textsuperscript{2}The Hong Kong University of Science and Technology
 \\
 \textsuperscript{3}China Mobile Information Technology Company Limited
\\
 \small{
    \href{mailto:email@domain}{tlin280@connect.hkust-gz.edu.cn}
 }
 \small{
    \href{mailto:email@domain}{\{yuyuluo,wuks,nantang\}@hkust-gz.edu.cn}
 }
 \\
 \small{
    \href{mailto:email@domain}{\{zhanghonglin, zhangjichengit, liuchunlin\}@chinamobile.com}
 }
}

\begin{document}
\maketitle

\begin{abstract}
Cross-Document Multi-entity Question Answering (MEQA) demands the integration of scattered information across documents to resolve complex queries involving entities, relationships, and contextual dependencies. Although Large Language Models (LLMs) and Retrieval-augmented Generation (RAG) systems show promise, their performance on cross-document MEQA remains underexplored due to the absence of tailored benchmarks. To address this gap, we introduce MEBench, a scalable multi-document, multi-entity benchmark designed to systematically evaluate LLMs’ capacity to retrieve, consolidate, and reason over scattered and dense information. Our benchmark comprises 4,780 questions which are systematically categorized into three primary categories: \textit{Comparative Reasoning, Statistical Reasoning} and \textit{Relational Reasoning}, further divided into eight distinct types, ensuring broad coverage of real-world multi-entity reasoning scenarios. Our experiments on state-of-the-art LLMs reveal critical limitations: even advanced models achieve only 59\% accuracy on MEBench. Our benchmark emphasizes the importance of completeness and factual precision of information extraction in MEQA tasks, using Entity-Attributed F1 (EA-F1) metric for granular evaluation of entity-level correctness and attribution validity. MEBench not only highlights systemic weaknesses in current LLM frameworks but also provides a foundation for advancing robust, entity-aware QA architectures.\footnote{The source code and data have been made available at \url{https://github.com/tl2309/MEBench}}
\end{abstract}
\section{Introduction}

\begin{figure}[t!]
\begin{center}
\includegraphics[width=1\linewidth]{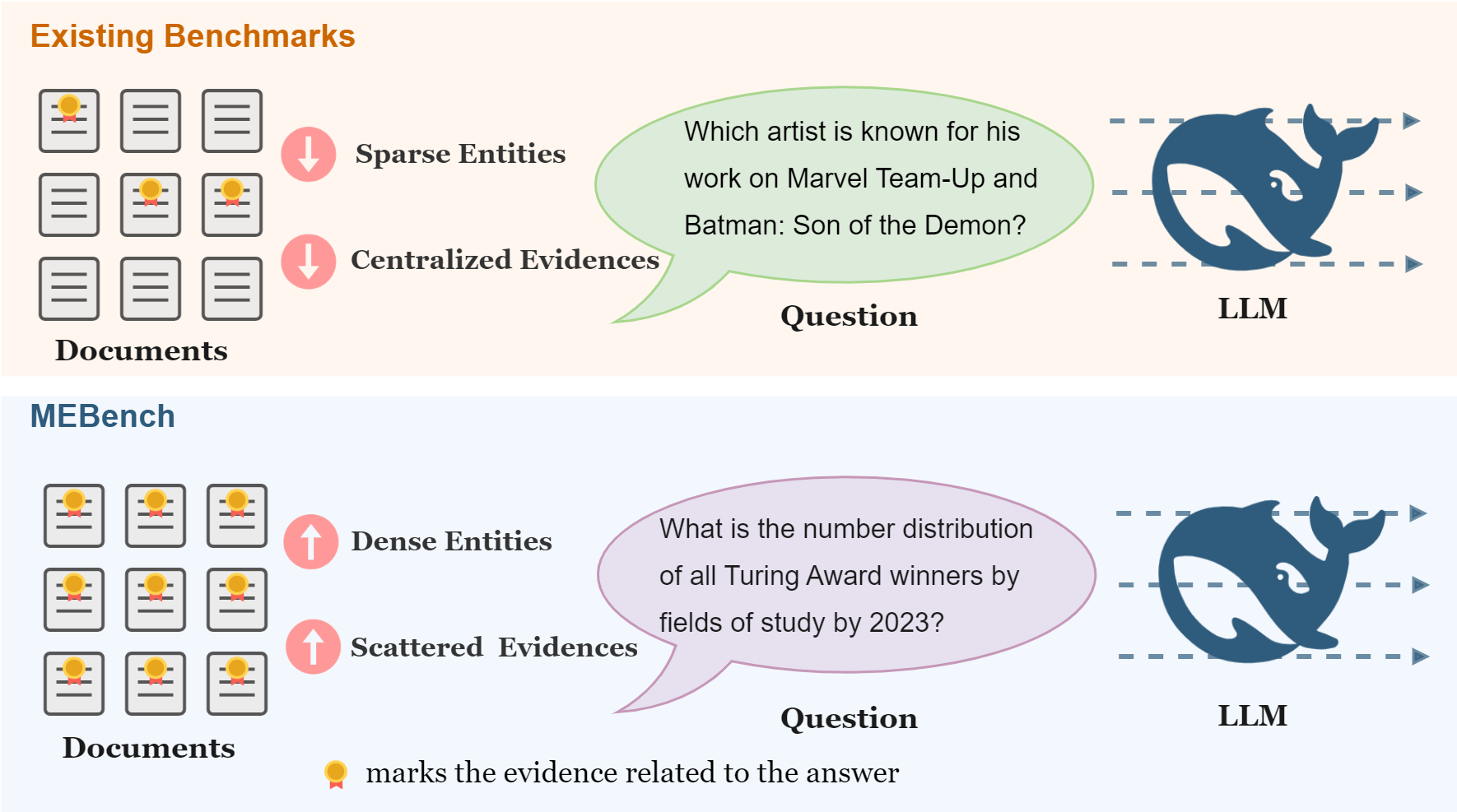}
\end{center}
\caption{Existing benchmarks vs. \ben. Unlike existing benchmarks which feature centralized evidence distributions and sparse entity mentions, \ben presents entity-dense scene where critical evidences are dispersed across multiple documents, necessitating that when seeking an answer, no document or entity can be ignored.}
\label{fig:benexa}
\end{figure}

The emergence of large language models (LLMs) has significantly advanced natural language processing capabilities, demonstrating exceptional performance in diverse tasks spanning text generation to data science and databases~\cite{achiam2023gpt,lin2025Simplifying,DBLP:journals/corr/abs-2504-01990,alphasql,aflow,nl2sql360,DBLP:journals/pacmmod/ChenFWTDWLTLZD25,DBLP:journals/pvldb/LiYLFT25,DBLP:conf/icde/FanHFC00024}. Nevertheless, long-context LLMs exhibit notable limitations in processing entity-dense analytical reasoning, particularly when contextual dependencies are distributed across multiple documents~\cite{DBLP:journals/corr/abs-2505-19716,DBLP:journals/corr/abs-2506-09507,shi2025trainabledynamicmasksparse}, and we analytically argue that context window limitations, over-reliance on parametric knowledge, and poor cross-document attention as the key bottlenecks~\cite{verifyai,statqa,DBLP:journals/corr/abs-2502-06855,DBLP:journals/corr/abs-2505-07437}. On the other hand, current implementations of retrieval-augmented generation (RAG) architectures~\cite{crag, wu2025clash,fan2024Asurvey,Tang2024graphgpt,liu2025surveytexttosqlerallms,mar, lin2025srag,11107459,DBLP:conf/acl/HongLLLWZLCZWZZ25} frameworks' effectiveness in addressing cross-document Multi-entity Question Answering (\meqa) remains insufficiently investigated. Furthermore, the field lacks comprehensive benchmarking frameworks specifically designed to evaluate the performance of LLMs and RAG systems for cross-document entity-intensive tasks. As shown in Figure~\ref{fig:benexa}, existing evaluation metrics frequently inadequately represent the complexities inherent in real-world MEQA applications~\citep{song2024counting}, where queries such as ``What is the number distribution of all Turing Award winners by fields of study by 2023?'' necessitate not only high-precision information retrieval but also reasoning over fragmented, entity-specific information across heterogeneous document sources.  

To address this methodological gap, we present MEbench, a novel benchmarking framework specifically designed to assess the performance of large language models and RAG systems in cross-document multi-entity question answering scenarios. The benchmark simulates real-world information integration challenges where correct answers require synthesizing entity-centric evidence distributed across multiple documents, with a single instance of document omission or entity misinterpretation can propagate errors through the reasoning chain. As shown in Table~\ref{tab:querycategories}, \ben features a mean entity density of 409 entities per query, with systematically varied entity cardinality across three operational tiers: low (0-10 entities), medium (10-100 entities), and high complexity (>100 entities). This stratified design enables granular performance evaluation across different entity scales and task difficulty levels. The framework comprises 4,780 validated question-answer pairs systematically categorized into three primary categories and eight distinct types, MEBench spans diverse real-world scenarios, from academic field distributions to geopolitical event analysis. Our experiments with state-of-the-art models, including GPT-4 and Llama-3, reveal significant shortcomings: even the most advanced LLMs achieve only 59\% accuracy on MEBench. This underscores systemic weaknesses in current frameworks, for example, models frequently fail to locate all entity and their attributes or infer implicit relationships, highlighting the need for architectures that prioritize entity-aware retrieval and contextual consolidation.

Our contributions are summarized as follows:  

\stitle{Development of MEBench.} A scalable benchmark designed to evaluate LLMs and RAG systems in cross-document aggregation and reasoning. It includes 4,780 validated question-answer pairs spanning three categories and eight types, simulating real-world scenarios that demand integration of scattered, entity-specific information.   

\stitle{Entity-centric Task Categories and Evaluation.} Utilization of Entity-Attributed F1 (EA-F1), a granular metric for assessing entity-level correctness and attribution validity, alongside a stratified entity density design (low: 0–10, medium: 11–100, high: >100 entities per query). Our framework emphasizes completeness and factual precision in information extraction, addressing gaps in existing metrics for entity-dense MEQA tasks.

\stitle{Scalable Benchmark Construction.} A scalable, automated pipeline: Knowledge graph extraction from structured Wikipedia for cross-document relationship discovery; Relational table generation to preserve entity-property relationships; Template-based QA generation ensuring reproducibility and reducing cost and labor.

\section{Related Work}

Recent advancements in question answering (QA) have been driven by breakthroughs in LLMs and RAG systems. While these technologies excel in single or a few document settings, demonstrating proficiency in tasks like fact extraction, summarization, and reasoning within a single source, their performance in cross-document, multi-entity scenarios remains underexplored. This section contextualizes our work within three key research areas: single-document QA, cross-document aggregation, and entity-centric evaluation.

\subsection{Single-Document QA and LLM Progress}

Many QA benchmarks, such as SQuAD~\cite{2016SQuAD}, Natural Questions ~\cite{2019Natural}, L-eval~\cite{an-etal-2024-l} and needle-in-a-haystack~\cite{Kamradt2023}, focus on extracting answers from individual document.  Modern LLMs like GPT-4~\cite{achiam2023gpt}, Llama-3~\cite{Llama3_2024}, and PaLM~\cite{2023PaLM} have achieved near-human performance on these tasks, leveraging their ability to parse and reason within localized contexts. However, these benchmarks do not address the complexities of integrating information across multiple documents, a critical limitation for real-world applications~\cite{DBLP:conf/icde/LuoQ0018,DBLP:conf/sigmod/LuoQ00W18,nvbench,ncnet,DBLP:journals/vldb/QinLTL20,DBLP:journals/corr/abs-2503-22402,DBLP:journals/corr/abs-2503-11984,DBLP:journals/corr/abs-2504-01990}.

\subsection{Cross-Document Aggregation Challenges}

Efforts to extend QA to multi-document settings include datasets like HotpotQA~\cite{2018HotpotQA}, MuSiQue~\cite{2021MuSiQue}, LooGLE~\cite{li2024loogle}, LM-Infinit~\cite{han-etal-2024-lm}, $\infty$ Bench~\cite{zhang2024inftybench}, CLongEval~\cite{qiu2024clong}, BAMBOO~\cite{dong-etal-2024-bamboo}, Loong~\cite{2024Leave} and Symphony~\cite{symphony}, which emphasize multi-hop reasoning and cross-source synthesis. While these benchmarks highlight the need for systems to connect disparate information, they often prioritize breadth over depth in entity-centric reasoning. For instance, questions in these datasets rarely demand the consolidation of attributes for dozens or more entities (\eg aggregating ACM Fellows' expertise across fields), a gap that limits their utility in evaluating entity-dense scenarios. Recent RAG frameworks~\cite{fan2024Asurvey, DBLP:journals/corr/abs-2504-10036} aim to enhance retrieval-augmented QA but struggle with ensuring completeness and attribution validity when handling multi-entity queries.

\subsection{Entity-Centric Evaluation Metrics.}
Existing evaluation metrics for QA, such as F1 score and exact match (EM), focus on answer surface-form correctness but overlook granular entity-level attribution~\cite{5707187}. Metrics in FEVER~\cite{thorne2018fever}, Attributed QA~\cite{bohnet2023attribut} and emphasize source verification, yet they lack the specificity to assess multi-entity integration. For example, they do not systematically measure whether all relevant entities are retrieved, their attributes are correctly extracted, or their sources are accurately used, which is a shortcoming that becomes critical in MEQA tasks.

\subsection{The Gap in Multi-Entity QA Benchmarks.}
Prior work has yet to establish a benchmark that systematically evaluates LLMs and RAG systems on entity-dense, cross-document reasoning. Current datasets either lack the scale and diversity of real-world multi-entity questions or fail to provide fine-grained metrics for assessing entity-level completeness and attribution~\cite{song2024counting, 2024Leave, bai2025longbenchv2}. MEBench addresses these limitations by introducing a comprehensive evaluation framework that challenges models to retrieve, consolidate, and reason over scattered entity-centric data across heterogeneous sources. By incorporating the Entity-Attributed F1 (EA-F1) metric, our benchmark advances the field toward more precise, entity-aware QA systems.


\section{\ben}

\begin{table*}[t!]
\centering
\renewcommand\arraystretch{1}
\caption{Examples of multi-entities queries.}
\vspace{.2em}
\begin{tabular}{clp{7.5cm}}
\toprule
\textbf{Categories} & \textbf{Types} & \textbf{Examples}  \\
\midrule
\multirow{2}{*}{Comparison} 
    & \multirow{1}{*}{Intercomparison}   & Which has more ACM fellow, UK or USA?\\
    \cmidrule(lr){2-3}
    & \multirow{1}{*}{Superlative} &Which city has the highest population?\\
\midrule
\multirow{8}{*}{Statistics} 
    &\multirow{1}{*}{Aggregation} &How many ACM fellow are from MIT?\\
    \cmidrule(lr){2-3} 
    &\multirow{2}{*}{Distribution Compliance} & Does the nationality of ACM fellows follow a normal distribution? \\
    \cmidrule(lr){2-3} 
    &\multirow{2}{*}{Correlation Analysis} & Is there a linear relationship between number of events and records broken in Olympic Games?\\
    \cmidrule(lr){2-3} 
    &\multirow{3}{*}{Variance Analysis} & Do the variances in the number of participating countries and total events in the Summer Olympics differ significantly?\\
\midrule
\multirow{4}{*}{Relationship} 
    & \multirow{2}{*}{Descriptive Relationship} &Is there a relationship between the year of ACM fellowship induction and the fellows' areas of expertise?\\
    \cmidrule(lr){2-3} 
    & \multirow{2}{*}{Hypothetical Scenarios} & If China wins one more gold medal, will it overtake the US in the gold medal tally at the 2024 Olympics?\\
\bottomrule
\end{tabular}
\label{tab:examples}
\end{table*}

\begin{table*}[t!]
  \centering
    \caption{Statistics of \ben benchmark.}
    \vspace{.2em}
  \begin{tabular}{lccc}
    \toprule
    \textbf{Categories}  & \textbf{\ben -train} & \textbf{\ben -test}& \textbf{\ben -total} \\
    \midrule
    \#-Queries  &3406 &1374 &4780 \\
     \#-Topics  &165 &76 & 241\\
     Ave. \#-entities /Q   &460 &391 & 409\\
     \hline
    \multicolumn{4}{c} {\textit{Hops}}\\
     \hline
    \#-one-hop Q    & 1406 &606 & 2012\\
    \#-multi-hop Q    & 1322& 768& 2090\\
    \hline
    \multicolumn{4}{c} {\textit{Categories}}\\
     \hline
    \#-Comparison  &1107 &438 & 1545\\
    \#-Statistics &1440 &585 & 2025\\
    \#-Relationship &859 &351 & 1210\\
    \hline
    \multicolumn{4}{c} {\textit{Entity Density}}\\
     \hline
    \#-low (0–10)  &487 &196 &683 \\
    \#-medium(11–100)  &973 &393 &1366 \\
    \#-high (>100)  &1946 &785 &2731 \\
   \bottomrule
  \end{tabular}
  \label{tab:querycategories}
\end{table*}

\subsection{Task overview}
\ben is a structured evaluation framework designed to systematically assess the capabilities of LLMs and RAG systems in performing cross-document multi-entity question answering. This framework targets three core reasoning modalities: comparative analysis, statistical inference, and relational reasoning, and each decomposed into specialized subtasks that rigorously test distinct facets of LLM performance, ensuring broad coverage of real-world multi-entity reasoning scenarios. Examples of tasks are provided in Table ~\ref{tab:examples}.
Each of three primary task categories addresses distinct reasoning challenges:

\paragraph{Comparative Reasoning}

Comparative reasoning tasks evaluate LLM’s ability to juxtapose entities across heterogeneous documents, demanding both attribute alignment and contextual synthesis.



\paragraph{Statistical Reasoning} 

Statistical tasks assess LLM’s proficiency in \textbf{quantitative synthesis}, including aggregation, distributional analysis, correlation analysis, and variance analysis across multi-document.





\paragraph{Relational Reasoning}
Relational tasks probe model’s capacity to model explicit interactions and counterfactual dependencies among entities.


\subsection{Benchmark Construction}
MEBench was constructed through a systematic pipeline, comprising the following steps.  

\begin{figure*}[t!]
\begin{center}
\includegraphics[width=1\linewidth]{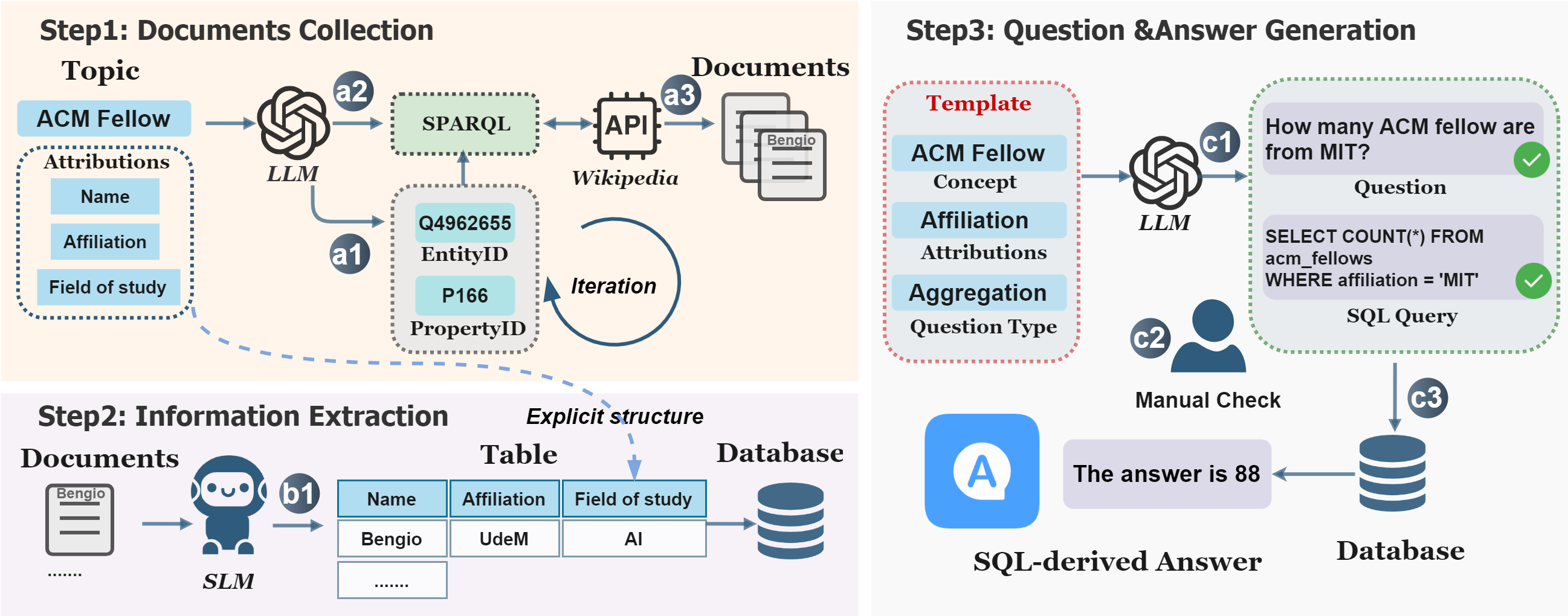}
\end{center}
  \caption{The systematic pipeline of benchmark construction. It comprising three phases: documents collection, information extraction and question-answer generation. In the documents collection phase, concept topics relevant to multi-entity scenarios are selected, followed by GPT-4 processing descriptions to extract entities and properties mapped to Wikipedia IDs for integration with structured Wiki data. Structured information from Wikipedia documents is processed using small language models (SLMs) due to the structured nature of the documents, culminating in table creation with entity attributes as columns. For QA generation, questions are generated following a "template-driven, entity-attribute coupling" paradigm using GPT-4 with predefined templates, and undergo syntactic, semantic, and ambiguity checks, while answers are programmatically derived via SQL queries against the table and standardized into canonical forms. The final dataset ensures traceability (SQL-derived answers), scalability (template-driven approach), and rigor (execution-based answering reduces hallucination risks).}
  \label{fig:mepipe}
\end{figure*}

\subsubsection{Data Collection} 

\paragraph{Concept Topic Identification.} In the initial phase of data collection for MEbench, a meticulous process is employed to determine the concept topics that are applicable to multi-entity scenarios. These topics are carefully selected based on their significance, prevalence, and the potential for generating complex multi-entity questions, and examples can be seen in Appendix Table~\ref{tab:topic-e}.

\paragraph{Entity and Property Identification.} Once the concept topics are determined, we input descriptions related to the concept topics into a LLM (we use GPT-4), which then processes the text to identify concept entity and property, as illustrated in Figure~\ref{fig:mepipe}-a1. After the LLM identifies the entity and property via iterative semantic refinement, we map them to entity IDs and property IDs in the Wiki graph. This mapping is crucial as it allows for seamless integration with the vast amount of structured data available in Wikipedia. The detailed method is in Appendix~\ref{sec:sparql}. Using the entity ID and property ID, we synthesise SPARQL. We then utilize the API provided by Wikipedia to retrieve the wiki web pages of all entities related to the topic. For example, if our concept topic is "ACM Fellows", we would obtain the Wikipedia pages of all ACM Fellows, which contain their detailed information. We use GPT-4 to generate a set of interesting entity attributes. These attributes are carefully chosen based on general interest and relevance in the domain. For ACM Fellows, as an example, nationality, research field, institution, and academic contribution maybe the attributes that people commonly pay attention to.

\paragraph{Structured Information Processing.} Once the document set is collected, we proceed to the structured information processing stage. The documents we have gathered from Wikipedia have well-defined and accurate structural relations. Due to the structured nature of the documents, we do not need to rely on the long context ability of large language models. Instead, we can use small language models (SLMs) for information extraction. They are well-suited for tasks where the information is already structured and the focus is on extracting specific details~\citep{fan2025minirag}.

\paragraph{Table Generation.} The final step in the data collection process is to generate a table, as shown in Figure~\ref{fig:mepipe}-b1. We use the the entity attributes as the column headers of the table. Each row in the table represents an individual entity. For example, in the case of ACM Fellows, each row would correspond to an individual ACM Fellow. 

\subsubsection{Question and answer Generation}  
The question and answer generation framework for \ben is a structured, multi-phase process that leverages LLM and tabular data to produce both semantically coherent questions and computationally verifiable answers.

\paragraph{Question Generation.}
The foundational input for the QA generation pipeline is the table generated in last step. The generation of questions follows a "template-driven, entity-attribute coupling" paradigm, implemented through LLM (GPT-4), as illustrated in Figure~\ref{fig:mepipe}-c1. Predefined syntactic and semantic templates govern the grammatical structure and intent of questions. These templates are shown in Appendix Table~\ref{tab:template}. The LLM instantiates templates with entity-attribute pairs, ensuring syntactic diversity while adhering to logical constraints. Generated questions undergo validation via: Syntactic checks, ensuring grammatical correctness; Semantic grounding, verifying that the question is answerable using the table’s data; Ambiguity reduction, pruning underspecified questions (\eg ``Describe the economy'' revised to ``Describe the GDP growth rate of Brazil in 2023'').
\paragraph{Answer Generation.}
Answers are derived programmatically through automated SQL query execution, ensuring reproducibility and alignment with the table’s ground-truth data. The synthesized SQL is executed against the table, yielding direct answers or sub-tables (Intermediate results requiring post-processing), as illustrated in Figure~\ref{fig:mepipe}-c3. Answers are standardized to ensure consistency: Numeric results are rounded to significant figures; Categorical answers are converted to canonical forms (\eg "USA" to "United States").

\subsection{Data Statistics}
The benchmark comprises 4,780 methodically structured questions partitioned into two subsets: a training set (3,406 questions) for model fine-tuning or train, and a test set (1,374 questions) for rigorous evaluation. Based on entity count, the data is divided into three groups: ``low'' (0-10), ``Medium'' (11-100), and ``high'' (>100), containing 683, 1366, and 2731 entries, respectively. Table~\ref{tab:querycategories} details comprehensive statistics of the benchmark. We also analyze the proportion of questions rejected during manual review and about 21\% of the questions are failure to meet quality standards.

\section{Experiment} 

\begin{table*}[t!]
  \centering
  \renewcommand\arraystretch{1}
  \caption{
   Experimental results for \ben. 
  }
  {\small
  \begin{tabular}{lcccc}
    \toprule
    \multirow{2}{*}{\textbf{Models}} 
        & \multicolumn{4}{c} {\textbf{Accuracy}}\\
        \cmidrule(lr){2-5}
        & Comparison &Statistics &Relationship &Overall\\
    \midrule
    \multicolumn{5}{c} {\textbf{All sets}}\\
     \hline
    GPT-3.5-turbo &0.105 	&0.198 	&0.476 	&0.239 \\
    GPT-3.5-turbo +  RAG & 0.605 	&0.260 	&0.476 	&0.425 \\
    GPT-4 & 0.199 	&0.289 	&0.507 	&0.316 \\
    GPT-4 + RAG & 0.763 	&0.410 	&0.687 	&0.593 \\
    Llama-3-Instruct & 0.046 	&0.118 	&0.256 	&0.130 \\
    Llama-3-Instruct + RAG & 0.447 	&0.181 	&0.410 	&0.325 \\
    FT Llama-3-Instruct & 0.046 	&0.253 	&0.259 	&0.189 \\
    FT Llama-3-Instruct + RAG & 0.687 	&0.448 	&0.573 	&0.556 \\
    \hline
    \multicolumn{5}{c} {\textbf{Set1 (0-10)}}\\
     \hline
      GPT-3.5-turbo &0.435	&0.583	&0.560	&0.530 \\
    GPT-3.5-turbo + RAG & 0.548 &0.654	&0.620	&0.612\\
    GPT-4 & 0.451	&0.595	&0.540	&0.535 \\
    GPT-4 + RAG & 0.870	&0.619	&0.740	&0.729 \\
    Llama-3-Instruct & 0.322 &0.500	&0.400	&0.418\\
    Llama-3-Instruct + RAG & 0.419 &0.571 &0.480	&0.500 \\
    FT Llama-3-Instruct & 0.322 &0.511 &0.380 &0.418\\
    FT Llama-3-Instruct + RAG & 0.580 &0.677 &0.690 &0.676 \\
    \hline
    \multicolumn{5}{c} {\textbf{Set2 (11-100)}}\\
     \hline
      GPT-3.5-turbo &0.364 	&0.495 	&0.544 	&0.466 \\
    GPT-3.5-turbo + RAG &0.613 	&0.581 	&0.640 	&0.607 \\
    GPT-4 & 0.348 	&0.476 	&0.521 	&0.447 \\
    GPT-4 + RAG & 0.791 	&0.511 	&0.661 	&0.638 \\
    Llama-3-Instruct & 0.240 	&0.385 & 0.357 	&0.332  \\
    Llama-3-Instruct + RAG & 0.428 	&0.454 	&0.459 	&0.447  \\
    FT Llama-3-Instruct & 0.240 	&0.434 	&0.344 	&0.349  \\
    FT Llama-3-Instruct + RAG & 0.612 	&0.608 	&0.655 	&0.640  \\
     \hline
    \multicolumn{5}{c} {\textbf{Set3 (>100)}}\\
    \hline
    GPT-3.5-turbo &0.09 &0.158 &0.291 &0.173\\
    GPT-3.5-turbo + RAG & 0.389 &0.191 &0.311 &0.285\\
    GPT-4 & 0.142 &0.202 &0.309 &0.210\\
    GPT-4 + RAG & 0.436 &0.270 &0.405 &0.357\\
    Llama-3-Instruct &0.055 &0.108 &0.168 &0.106\\
    Llama-3-Instruct + RAG & 0.265 &0.147 &0.253 &0.212\\
    FT Llama-3-Instruct & 0.055 &0.177 &0.167 &0.136\\
    FT Llama-3-Instruct + RAG & 0.401 &0.291 &0.355 &0.345\\
    \bottomrule
  \end{tabular}
  }
  \label{tab:Accuracy}
\end{table*}

\begin{table}[t!]
\caption{\label{tab:f1}
	Quality of Large Language Models (LLMs) in EA-F1. 
	}
	\centering
	\begin{tabular}{lc}
		\hline
		\multirow{2}*{\textbf{Models}} & \multirow{2}*{\textbf{$EA-F1$}}  \\
        &\\
		\hline
		GPT-3.5-turbo & 0.25  \\
		GPT-3.5-turbo + RAG & 0.43\\
		GPT-4 & 0.36 \\
		GPT-4 + RAG & 0.71 \\
		Llama-3-Instruct & 0.21 \\
		Llama-3-Instruct + RAG & 0.39 \\
            FT Llama-3-Instruct & 0.21 \\
		FT Llama-3-Instruct + RAG & 0.59 \\
		\hline
	\end{tabular}
\end{table}

\begin{figure*}[t!]
\begin{center}
\includegraphics[width=1\linewidth]{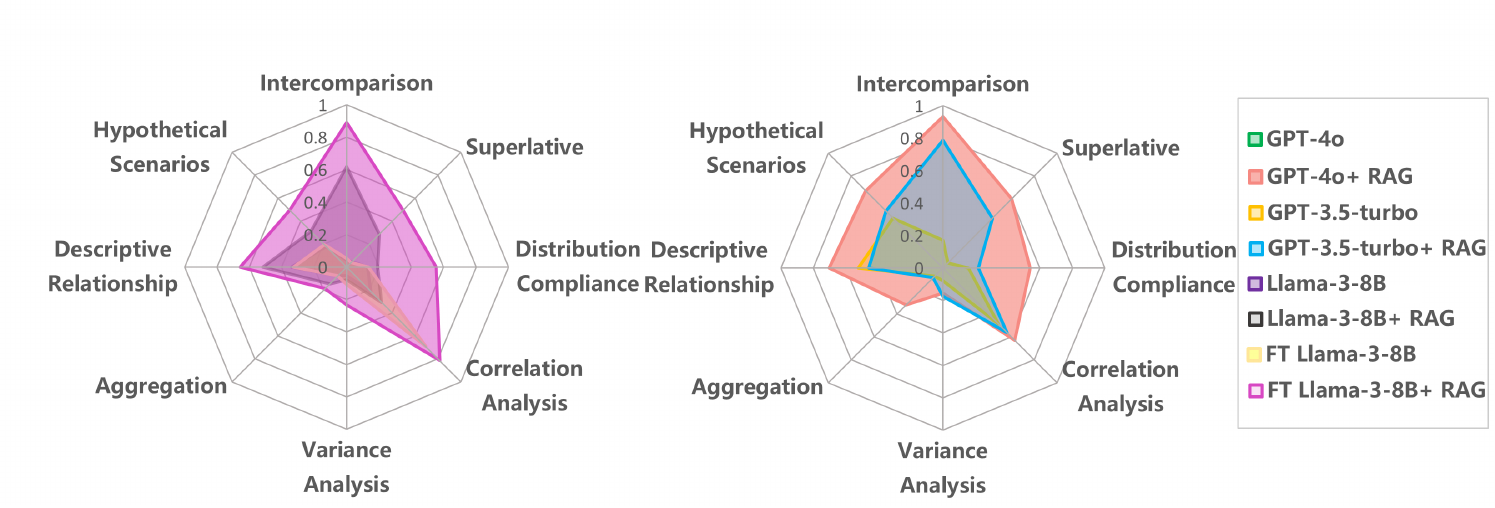}
\end{center}
  \caption{The Experimental results for eight subtasks of each model.}
  \label{fig:acc}
\end{figure*}

\subsection{Experiment Setup}
\stitle{Models.}
For open-source LLMs, we conduct experiments using the representative Meta-Llama-3-8B-Instruct~\citep{Llama3_2024} and apply QLoRA~\citep{dettmers2023qlora} to fine-tune it with the training set of \ben. For proprietary LLMs, we select the widely recognized GPT models, including GPT-3.5-turbo~\citep{ouyang2022training} and GPT-4~\citep{achiam2023gpt}. 

\stitle{RAG.}
We implement a hierarchical retrieval framework that explicitly incorporates document organizational structures into the RAG pipeline to explore whether RAG can enhance the model’s performance on \ben. For the Embedding choice, we employ the OpenAI Embedding model~\citep{TextEmbeddingAda002}, and the chunk size is 1024. For each document, we retrieve the top-5 most related chunks and concatenate them in their original order to form the context input for the model.
 
\stitle{Evaluation Metrics.}
We adopt Accuracy ($Acc$) as the primary metric to assess the performance of LLMs on \ben tasks. For the subcategories of Variance Analysis, Correlation Analysis, and Distribution Compliance within the Statistics tasks, which are shown in Table~\ref{tab:examples}, we focus solely on prompting LLMs to identify relevant columns and applicable methods, evaluating the accuracy of their selections instead of the computational results, as LLMs' abilities in precise calculations are not the central focus of this study. In addition, we evaluate performance of information extraction using Entity-Attributed F1 (EA-F1). This is an F1 score applied to the predicted vs. gold sets of the \colorbox{black!5}{(entity, atrribution, value)}. All three elements in the tuple must exactly match the tuple in the ground truth to be marked correct.  

\subsection{Results and Analysis}

Various models exhibit notable variations in performance on \ben. 
Table~\ref{tab:Accuracy} presents experimental results alongside overall accuracy on \ben, and Figure~\ref{fig:acc} shows accuracy on eight further-divided tasks.

\stitle{Main result.}
GPT-4 + RAG achieved superior accuracy (59.3\%), outperforming the second-ranked model (FT Llama-3-Instruct: 55.6\% ) by a statistically significant margin. Notably, GPT-4 + RAG excelled in relational (68.7\%) and comparative (76.3\%) queries, likely due to its superior contextual understanding. However, all models exhibited markedly lower accuracy in statistical queries (GPT-4 + RAG: 41.0\%), suggesting inherent challenges in numerical reasoning. In our evaluation, we focused on analyzing the capability of LLMs to extract question-related data. This assessment aimed to understand how well these sophisticated models can organize and present data for the question. The result is shown in Table~\ref{tab:f1}. These results underscore the critical role of information extraction architectures in mitigating hallucinations and grounding outputs in factual data. 
Introducing RAG significantly improves overall performance, particularly in comparison tasks, while fine-tuning LLaMA-3-Instruct alone does not yield substantial gains without RAG. On \ben, open-source models like LLaMA-3-Instruct, even with RAG, can't match proprietary models like GPT-4, which achieves a 59.3\% accuracy compared to LLaMA-3-Instruct's 32.5\%.

\stitle{Fine-grained Performance on Sub-tasks.}
Figure~\ref{fig:acc} shows that vanilla LLMs perform well in correlation analysis and descriptive relationship tasks, while RAG significantly improves intercomparison and superlative tasks. However, neither fine-tuning nor RAG overcomes challenges in variance analysis and aggregation tasks, while GPT-4 + RAG achieves accuracy of 15.3\% and 32.1\%.

\stitle{Entity density Analysis.} 
As we can see from Table~\ref{tab:Accuracy}, our experiments underscore the impact of entity density on model performance in MEQA tasks. This phenomenon arises because higher entity densities amplify two critical challenges inherent to MEQA systems: (1) Semantic ambiguity due to overlapping relational predicates among entities (e.g., distinguishing "Paris [person]" vs. "Paris [location]" within narrow contexts), and (2) computational overhead in attention-based architectures attempting parallel reasoning over entangled entity-attribution pairs (e.g. transformer self-attention weights saturate under dense cross-entity dependencies). 

\begin{itemize}
 \item {Low Entity Density}: Models generally performed well in low-density scenarios. The simplicity of context allowed for accurate entity recognition and minimal ambiguity.
 \item {Medium Entity Density}: Performance began to decrise among models in medium-density scenarios by 6\% average acc. This variance suggests differences in how models handle increased entity complexity and overlapping contexts.
 \item {High Entity Density}: High-density questions posed significant challenges, with an average acc drop to 22.8\% across models. The result highlighting limitations in current architectures' ability to handle complex multi-entity questions. 
\end{itemize}

\section{Limitations} 
While MEBench provides a comprehensive framework for evaluating cross-document multi-entity reasoning, our work has several limitations that warrant further investigation. Although MEBench covers eight distinct reasoning types across three broad categories, real-world MEQA scenarios may involve even more intricate combinations of logical, temporal, or causal dependencies. The current benchmark does not explicitly model dynamic or time-sensitive entity interactions, which could limit its applicability to domains like financial forecasting or event-driven narratives. The benchmark relies on a curated collection of documents to ensure controlled evaluation. While this design choice minimizes noise, it may not fully replicate the challenges of real-world environments where documents vary widely in quality, redundancy, and structure. Future iterations could incorporate noisy or incomplete data sources to better simulate practical scenarios. While the Entity-Attributed F1 (EA-F1) metric rigorously assesses entity-level correctness and attribution validity, it prioritizes factual precision over semantic coherence. This may undervalue partially correct answers that demonstrate valid reasoning chains but contain minor factual inaccuracies. A hybrid evaluation framework combining EA-F1 with human judgment could provide a more holistic assessment.
\section{Conclusion} 
In this study, we have comprehensively addressed the significant challenges that Multi-entity Question Answering (MEQA) poses to LLMs and RAG systems. The limitations of existing methods in handling cross-document aggregation, especially when dealing with entity-dense questions, have been clearly identified and analyzed. We introduced MEBench, a groundbreaking multi-document, multi-entity benchmark. Our experiments on state-of-the-art LLMs such as GPT-4 and Llama-3, along with RAG pipelines, have shed light on the critical limitations of these advanced models. The fact that even these leading models achieve only 59\% accuracy on MEBench underscores the magnitude of the challenges in MEQA. MEBench has effectively highlighted the systemic weaknesses in current LLM frameworks. These weaknesses serve as valuable insights for future research directions. For instance, the need for improved algorithms to retrieve and consolidate fragmented information from heterogeneous sources is evident. Additionally, there is a pressing need to develop more robust entity-aware QA architectures that can better handle the complexities of MEQA.

\section{Acknowledgment} 

This work is supported by Guangdong provincial project 2023CX10X008.


\bibliography{refs/custom}

\clearpage
\appendix

\section{Appendix}
\label{sec:appendix}

\begin{table*}[t!]
\centering
\caption{
Example Topics and Their Entities Attributions. 
}
\vspace{.2em}
\begin{tabular}{lp{7cm}c}
\toprule
    \textbf{Topics}           & \textbf{Entities Attributions}  & \textbf{\#-Entities}\\
    \midrule
    \multirow{1}{*}{ACM fellow} & nationality, field of study, affiliation & \multirow{1}{*}{1115}\\
    \hline
    \multirow{2}{*}{Presidents of the US} & term lengths, political parties, vice-presidents, birth states, previous occupations & \multirow{2}{*}{55}\\
    \hline
    \multirow{2}{*}{Chemical Elements} &  atomic number, atomic mass, boiling point, melting point, electron configuration & \multirow{2}{*}{166}\\
    \hline
    \multirow{2}{*}{Summer Olympic Games} & host cities, number of participating countries, total number of events, medal tally, records broken & \multirow{3}{*}{35}\\
    \hline
    \multirow{2}{*}{Nobel Prize in Chemistry} &  categories, year of award, country of origin, field of contribution.& \multirow{2}{*}{194}\\
    \hline
    \multirow{1}{*}{Cities of the World} &  population, geographic coordinates, altitude, GDP  &\multirow{1}{*}{7040}\\
\bottomrule
\end{tabular}
\label{tab:topic-e}
\end{table*}

\begin{table*}[t!]
\centering
\renewcommand\arraystretch{1}
\caption{Template example for questions generated by the LLM (GPT-4).}
\vspace{.2em}
\begin{tabular}{clp{7.5cm}}
\toprule
\textbf{Types} & \textbf{Sub-types} & \textbf{Template Examples}  \\
\midrule
\multirow{2}{*}{Comparison} 
    & \multirow{1}{*}{Intercomparison}   & Which has high [property], [entity A] or [entity B]?\\
    \cmidrule(lr){2-3}
    & \multirow{1}{*}{Superlative} & Which [entity] has the highest/lowest [property]?\\
\midrule
\multirow{6}{*}{Statistics} 
    &\multirow{1}{*}{Aggregation} & How many [entities] have [specific property value]?\\
    \cmidrule(lr){2-3} 
    &\multirow{1}{*}{Distribution Compliance} & Does [property] follow a normal distribution? \\
    \cmidrule(lr){2-3} 
    &\multirow{2}{*}{Correlation Analysis} & Is there a linear relationship between [property A] and [property B]?\\
    \cmidrule(lr){2-3} 
    &\multirow{2}{*}{Variance Analysis} & Are the variances in [property A] and [property B] significantly different?\\
\midrule
\multirow{3}{*}{Relationship} 
    & \multirow{1}{*}{Descriptive Relationship} & How is [entity A] related to [entity B]?\\
    \cmidrule(lr){2-3} 
    & \multirow{2}{*}{Hypothetical Scenarios} & What would be the impact if [entity A] collaborates with [entity B]?\\
\bottomrule
\end{tabular}
\label{tab:template}
\end{table*}

\subsection{Methodology for composite SPARQL Generation via Iterative Semantic Refinement}\label{sec:sparql}  
\subsubsection{Initial Query Parsing Using GPT-4} We employ a transformer-based large language model (LLM), specifically GPT-4, to perform preliminary natural language question decomposition. This stage generates a proto-SPARQL query containing candidate triple patterns with hypothesized entity-property relationships. While this initial output captures broad syntactic structures (e.g., basic graph pattern groupings), it frequently exhibits two critical inaccuracies:

Entity Misalignment: Incorrect Wikidata Q-ID assignments due to lexical ambiguity (e.g., "Java" as programming language vs. Indonesian island)

Property Mismatch: Invalid P-ID selections arising from underspecified predicate semantics (e.g., using P19 [place of birth] instead of P20 [place of death])

\subsubsection{Semantic Validation Layer} To address these limitations, we implement a multi-stage correction framework:

(a) Structured Knowledge Anchoring

The system interfaces with the Wikipedia API through programmatic endpoints that map surface forms to canonical entities.



        
    
    
(b) Neural-Semantic Disambiguation Module

GPT-4 serves as our semantic analysis engine, performing three key operations: 

\begin{itemize}
 \item Contextual disambiguation using entity linking algorithms enhanced by Wikifier-style mention detection.
 \item Property type validation against Wikidata's ontology constraints (rdf:type, owl:ObjectProperty).
 \item Temporal scope verification for time-sensitive queries requiring qualifiers like P585 [point in time]. 
\end{itemize}

\subsubsection{Iterative Refinement Protocol} The system implements closed-loop feedback through successive cycles of:

\begin{itemize}
 \item Executing candidate SPARQL on the Wikidata Query Service endpoint.
 \item Analyzing result cardinality and type consistency.
 \item Applying constraint satisfaction heuristics: 
\end{itemize}

\begin{mdframed}
FILTER (?population > 1e6 \&\& ?country IN wd:Q30) \# Example numerical/entity constraints 
\end{mdframed}
Each iteration tightens precision metrics until meeting termination criteria defined by either:

$\frac{|ValidResults_t|}{|TotalResults_t|} \geq \theta_{precision} \quad $

$(\theta = 0.98\;\text{empirically})$

or maximum iteration thresholds.

\subsubsection{Final Query Synthesis} Through combining LLM-based semantic parsing with knowledge-grounded verification, we converge on an optimized SPARQL template satisfying both syntactic validity and functional correctness requirements for structured knowledge extraction.
\subsection{Optimization} 
Two aspects of optimization are included in \ben system to enhance the overall performance:


\paragraph{Model Selection.} \label{sec:model_selection} 
Model selection is straightforward yet highly effective for optimization~\citet{liu2024declarative}. Our system comprises multiple tasks, necessitating the selection of the most suitable model for different tasks. For basic tasks, more affordable and faster LLMs can suffice, while utilization of the most advanced LLMs is essential in more complex tasks to ensure optimal performance. Specifically, our system employs powerful yet resource-intensive GPT-4 for tasks such as semantic analysis or generation of table schemas and SQL queries. In contrast, for more basic information extraction, we utilize open-source Mistral-7B, thereby achieving a balance between cost efficiency and functional performance.

\paragraph{LLM Input/Output Control} SplitWise~\citet{patel2023splitwise} shows that LLM inference time is generally proportional to the size of input and output tokens. Since GPT models decide the cost based on the input token, we try to minimize the input of large models. Meanwhile, we use the instructive prompt to reduce the size of the outputs generated by LLM without changing the quality of these outputs. The example of prompt is in Appendix~\ref{appendix:output}.

\subsubsection{Prompt for Output Control}\label{appendix:output}
\begin{mdframed}
Review your output to ensure it meets all the above criteria. Your goal is to produce a clear, accurate, and well-structured output. Just output the result, no other word or symbol.
\end{mdframed}

\subsubsection{Quality Control} 
We devise several strategies to ensure the integrity and effectiveness of questions.

\paragraph{Question Templates.}
The use of templates ensures that every question is crafted with a clear structure, making it easier for respondents to understand and answer them accurately. For relationship and complex statistic questions we turn the questions in a closed-ended style, as they require a specific response of either "yes" or "no", which make the answer in a standardized format. The examples of Question Templates is in the Appendix \ref{tab:template}. 

\paragraph{Question Refinement.} After initial development, each question undergoes a refinement process which we used GPT-3.5-Turbo. This stage is critical for enhancing the clarity, relevance, and neutrality of the questions. It involves reviewing the questions for bias. This strategy helps in reducing misunderstandings and improving the overall quality of the questions.

\paragraph{Manual review.} We assess the questions for accuracy, ensuring they are factually correct and relevant to our purpose. Manual reviews can also provide insights into whether the questions are likely to effectively elicit the intended information from answers, thereby contributing to the reliability and validity of the benchmark.



\subsection{Tables}
Table~\ref{tab:topic-e} shows examples of topics and their entities' attributions. Table~\ref{tab:template} shows examples of question templates to synthesize questions.

\end{document}